\documentclass[journal]{IEEEtran}
\makeatletter
\def\endthebibliography{%
	\def\@noitemerr{\@latex@warning{Empty `thebibliography' environment}}%
	\endlist
}
\makeatletter

\ifCLASSINFOpdf
\else
\fi
\usepackage{graphicx}

\usepackage[cmex10]{amsmath}
\usepackage{algorithmic}

\usepackage{array}

\usepackage{multirow}

\usepackage{capt-of}

\usepackage{fixltx2e}

\usepackage{interval}

\usepackage{stfloats}
\begin{document}
%

\IEEEoverridecommandlockouts
\IEEEpubid{\begin{minipage}{\textwidth}\ \\[11pt] \centering
		\copyright 2019 IEEE. Personal use of this material is permitted.  Permission from IEEE must be including\\ reprinting/republishing this material for advertising or promotional purposes, creating new collective works,\\ for resale or redistribution to servers or lists, or reuse of any copyrighted component of this work in other works.
\end{minipage}} 
\title{Predicting Student Performance in an Educational Game Using a Hidden Markov Model}
\IEEEpubidadjcol
%

\author{Manie~Tadayon   ~~~~~~~~  Greg~Pottie
\thanks{Both authors are with the Electrical and Computer Engineering Department at University of California Los Angeles (UCLA).
	
Email: manitadayon@ucla.edu, pottie@ee.ucla.edu.}}

\maketitle

\begin{abstract}
$\textit{Contributions}$: Prior studies on education have mostly followed the model of  the cross sectional study, namely, examining the pretest and the posttest scores. This paper shows that students' knowledge throughout the intervention can be estimated by time series analysis using a hidden Markov model.

$\textit{Background}$: Analyzing time series and the interaction between the students and the game data can result in valuable information that cannot be gained by only cross sectional studies of the exams.

$\textit{Research Questions}$: Can a hidden Markov model be used to analyze the educational games? Can a hidden Markov model be used to make a prediction of the students' performance?

$\textit{Methodology}$: The study was conducted on (N=854) students who played the Save Patch game. Students were divided into class 1 and class 2. Class 1 students are those who scored lower in the test than class 2 students. The analysis is done by choosing various features of the game as the observations.

$\textit{Findings}$: The state trajectories can predict the students' performance accurately for both class 1 and class 2. 
\end{abstract}

\begin{IEEEkeywords}
education, game, hidden Markov model, prediction, time series.
\end{IEEEkeywords}

%

\section{Introduction}
%
%
%
%
\IEEEPARstart{E}{ducational} video games have received much attention in recent years due to their positive impacts on students' learning and their cognitive skills \cite{chuang2007effect}. However, just  because a game has educational content and is engaging does not mean it will be effective \cite{fisch2005making}. To prove its effectiveness, it needs to be further tested and analyzed. Fortunately every action, time click, and interaction in the game can be recorded. This provides a good opportunity for researchers to design a more sophisticated model and build more intelligent platforms.

Time series prediction has a rich history in domains such as speech processing, the stock market, and weather forecasting. Methods have been developed to perform robust and reliable forecasting using various machine learning and optimization algorithms \cite{de200625}, \cite{brockwell2002introduction}. 

The hidden Markov model (HMM) is a popular method to model the time series data because of its rich mathematical structure and the availability of many practical algorithms for computing model components \cite{rabiner1989tutorial}. Numerous papers such as \cite{rabiner1989tutorial}, \cite{varga1990hidden}, \cite{schuller2003hidden}, \cite{hassan2005stock}, \cite{durbin1998biological} about the HMM applications in speech, the stock market, and biology have been published; however there is a limited amount of work done in predicting the player's strategies or actions in a game using the HMM. For example, in  \cite{xie2002structure} the authors incorporated a two state HMM along with dynamic programming to classify and segment a soccer video game. In \cite{bunian2018modeling} the authors used a five state HMM to analyze the individual differences in game behavior and used the logistic regression for the prediction. They showed that the HMM based prediction using  sequential data gives better accuracy than a prediction using the aggregated data.
Some work has been done on modeling video games using dynamic Bayesian networks (DBN), such as \cite{huang2006semantic} and \cite{wang2005generic}. They focus on semantic analysis of sport video games. Considerable research, e.g. \cite{carmona2008designing}, \cite{gamboa2002designing}, \cite{conati1997line}, \cite{piech2015deep}, \cite{corbett1994knowledge}, and \cite{yudelson2013individualized} has been conducted on student modeling and designing intelligent tutoring systems (ITS) using Bayesian and belief networks. In \cite{corbett1994knowledge} and \cite{yudelson2013individualized} the authors used the Bayesian knowledge tracing (BKT) to model and evaluate student performance. BKT is a two state HMM where the probability of forgetting a skill is set to zero. However, to the best of our knowledge this is the first work that analyzes student performance in educational video games using an HMM.
  
The contribution of this paper is to present a novel approach to predict student performance using a video game as opposed to the exam. 

The rest of this paper is organized as follows. Section II reviews the HMM algorithm. Section III describes the game dataset used in this paper. Section IV describes the problem formulation as well as the prediction methods. Section V presents and discusses the results. Section VI concludes the paper and suggests a future work.


\section{HMM Algorithm}
In this section, HMM algorithms are briefly reviewed. Both the discrete hidden Markov model (DHMM) as well as the continuous hidden Markov model (CHMM) are discussed.
\IEEEpubidadjcol 
The HMM is the extension of the Markov process in which the observations are a probabilistic function of the states. In an HMM, states are considered as hidden and should be inferred by the sequence of observations. 

The HMM is characterized by the following: 

N: Number of the hidden states. Although this is unknown since the states are hidden, it usually can be initialized to a reasonable number depending on the problem and the dataset and later can be learned using various statistical analysis tools which will be discussed later. 

M: Number of the observation symbols per state. 

$\overrightarrow{S}$: State sequence where $\overrightarrow{S}=(s_1,s_2,...s_T)$, T is the length of the sequence, and each $s_i\in \{1,2,...,N\}$.

$\overrightarrow{O}$: sequence of the observation symbols where $\overrightarrow{O}=(o_1,o_2,...o_T)$ and each $o_i\in \{1,2,...,M\}$.

A: State transition probability. It defines the probability of going from state i to the state j and is denoted by
\begin{align}
a_{ij}=p(s_{t+1}=j|s_t=i)
\end{align}

B: Observation distribution per each state, which is denoted as follows:
\begin{align}
b_{i}(k)=p(o_{t}=k|s_t=i)
\end{align}

$\overrightarrow{\pi}$: Initial state distribution that is defined as follows:
\begin{align}
\overrightarrow{\pi_i}=p(s_1=i)
\end{align}

$\lambda$: HMM parameters together are usually denoted by the following:
\begin{align}
\lambda=(A,B,\overrightarrow{\pi})
\end{align}
The above equations together can be used to fully define any HMM with discrete observations. 

Forward and backward algorithms \cite{rabiner1989tutorial} are used to calculate $P(\overrightarrow{O}|\lambda)$, the probability of observing a sequence given $\lambda$. If the time series is not labeled and the mapping between the observations and the states is not available, then HMM parameters should be estimated using the Baum-Welch or EM algorithm \cite{dempster1977maximum}. If the observations are continuous (CHMM) as opposed to discrete, the emission probability distribution should be adjusted to account for this change. Continuous observations are modeled by fitting the probability density functions (pdf) to the data. A Gaussian distribution or mixture of Gaussian distributions are typically used for modeling the data. 

If the observations for each state can be modeled using a single Gaussian distribution, then equation (2) will be changed to the following:
\begin{align}
b_i(x)=p(x|s_t=i)=\textit{N}(x;\mu_i,\Sigma_i)
\end{align}
In equation (5), $\mu_i$ and $\Sigma_i$ are the mean and the covariance matrix of the Gaussian distribution for state i respectively.
     
If a single distribution is not a reasonable fit to the data, then a mixture of Gaussian distributions can be used to model the observations. In this case, equation (6) can be used to model the observations for each state.
\begin{align}
b_i(x)=p(x|s_t=i)=\sum_{m=1}^{M}c_{im}\textit{N}(x;\mu_{im},\Sigma_{im})
\end{align}
$c_{im}$ is the mixture coefficient and determines the weight each component has in modeling the data. $\mu_{im}$ and $\Sigma_{im}$ are the mean and covariance matrices of each mixture component corresponding to the state i. 

Decoding the optimal state sequence given the observation can be done using the Viterbi algorithm \cite{forney1973viterbi}. It finds the sequence of the states that best explains the observed data: 
\begin{align}
S^*=\textit{argmax}_S P(S|O,\lambda)
\end{align}
\section{Dataset}
The dataset used in this paper belongs to the Save Patch (SP) game designed by the National Center for Research on Evaluation, Standards, and Student Testing (CRESST). This game is one out of four fraction games designed to teach the concept of a unit in rational numbers. It is intended to teach the following two concepts: 1- Rational numbers are defined relative to a whole unit; 2- Rational numbers can be added only if they have a common denominator \cite{chung2012primer}.

Along with the four games, a pretest and a posttest were designed to test the students' understanding of the concepts before and after each game. A set of the questions targeted by each game in the posttest and pretest is carefully identified. This is very beneficial since it permits each game to be analyzed and verified independently of all the other games. 
\section{Problem Formulation}
In this section, the problem formulation and the prediction algorithm using the HMM are discussed. Prediction begins by dividing the students according to their score in the SP game into two classes: Class 1 are those who score low in the questions targeted by the SP game and Class 2 are those who score high in those questions. Each class is trained separately using the HMM, and the optimal parameters are determined by the model selection algorithms, which will be discussed later in the section. Testing or decoding is done by running the Viterbi algorithm on the observation sequences. 
 
Akaike Information Criterion (AIC) and Bayesian Information Criterion (BIC) are the model selection algorithms that are used to combat overfitting by introducing the penalty terms. AIC is defined by the following formula:
\begin{align}
AIC= -2\ln L+2k
\end{align}
where $\ln L$ is the log likelihood function and k is the number of parameters in the model; therefore 2k is the penalty term.

BIC is another well known model selection algorithm that measures the trade off between the model fit and the complexity. The formula for the BIC is given below:
\begin{align}
BIC= -2\ln L+k\ln N
\end{align}
$\ln L$ and k are the same parameters as AIC, and $N$ is the number of observations. By comparing equations (8) and (9) it appears that BIC has a larger penalizing term. Therefore it penalizes the complex model more than AIC. 

The prediction problem using the HMM is solved as follows:
Hidden states are defined to be the students' mastery levels, and the goal is to predict their final mastery level as they go through different levels of the game. This can be formulated as predicting the final mastery level $\hat{S}$ given all the past mastery levels $(s_1,s_2,...s_n)$.

The following techniques can be used to perform the prediction.
\begin{enumerate}
	\item \textbf{Naive}: This is the most basic method in which the predicted value is simply equal to the last observed value of the time series. 
	\begin{align}
	\hat{S}=s_n
	\end{align}	
	\item \textbf{Linear averaging}: This means that the final predicted value is the average of all the other mastery levels. 
	\begin{align}
	\hat{S}=\frac{\sum_{i=1}^{i=n}s_i}{n}
	\end{align}	
	One extension to this method is to perform averaging over a window of time length p, which means to consider only the most recent p values.
	\begin{align}
	\hat{S}=\frac{\sum_{i=n-p+1}^{i=n}s_i}{p}
	\end{align}
	Another extension of this method is the exponential smoothing. The idea is to perform the linear averaging by choosing  larger weights for the most recent values and smaller weights for the distant values. This is described by the following formula:
	\begin{align}
	\hat{S}=\alpha s_n+ \alpha (1-\alpha) s_{n-1} + \alpha (1-\alpha)^2 s_{n-2} + ...\notag
	\end{align}
	\begin{align}
	0\leq \alpha\leq 1 &
	\end{align}
	Equation(13) can also be written recursively as follows:
	\begin{align}
	&\hat{S}_n=\alpha s_n + (1-\alpha)\hat{S}_{n-1}\notag\\      
	&0\leq \alpha\leq 1 
	\end{align}
		\item \textbf{Mode}: This means that the final mastery level is the mastery level that appears most in the sequence. Mathematically this can be represented as follows:
	\begin{align}
	\hat{S}=\operatorname*{argmax}_j \sum_{i=1}^{i=n} 1 (S_i=j)
	\end{align}
\end{enumerate}
$\alpha$ in the equation (13) and (14) is referred to as the smoothing constant.  It is a parameter and is selected based on how important are the past values compared to the more recent values in a given time series. For example Table I shows the single exponential smoothing coefficients for the five most recent values in a time series. 

\begin{table}[!t]
\caption{Smoothing Coefficients for various $\alpha$}

\vspace{-1em}
  \begin{center}
	\begin{tabular}{c|c|c|c|c} 
		\hline
		 & $\alpha = 0.05$  & $\alpha = 0.1$    & $\alpha = 0.5$ & $\alpha = 0.9$\\  
		\hline
		$s_n$ & 0.05 & 0.1 & 0.5 & 0.9  \\ 
		\hline
		$s_{n-1}$ &0.0475 & 0.09 & 0.25 & 0.09  \\
		\hline
		$s_{n-2}$ & 0.0451 & 0. 0.081 & 0.125 & 0.009  \\
		\hline
		$s_{n-3}$ & 0.0429 & 0.0729 & 0.0625 & 0.0009  \\
		\hline
		$s_{n-4}$ & 0.0407 & 0.06561& 0.03125 & 0.00009 \\  
		\hline
	\label{TB1}	
	\end{tabular}
  \end{center}

\end{table}
As Table \ref{TB1} shows, a smaller value of $\alpha$ puts more weight on the more distant values and larger $\alpha$ put more weight on the more recent values of a time series.
\section{Results and Discussions} 
In this section, the results for the prediction task described in the last section are presented and discussed. The prediction is done by the DHMM by discretizing the observations to a certain number of bins using the domain knowledge or the kmean algorithm \cite{macqueen1967some}. Since the HMM training is done using expectation maximization (EM), and EM might converge to the local optimum instead of the global optimum, multiple initial conditions are used to test the algorithms and the best model is selected using the  AIC or BIC. The prediction is performed under the following cases:
\begin{enumerate}
	\item The total number of attempts per level is used as the observations.
	\item The total number of moves per level is used as the observations.
\end{enumerate} 
$\textbf{Case 1}$: Total number of attempts per level is used as the observation. 

The observations are discretized (DHMM) to four levels according to the following rules: 1 or 2 attempts per level is label 1; 3 or 4 attempts is label 2; 5,6 or 7 attempts is label 3; and anything above is label 4. According to the Tables \ref{TB2} and \ref{TB3} that provide the results for the model selections using the BIC algorithm, training is done by assigning the number of states (Q) to be 3 with the initial condition to be 35 for the class 1 and Q=2 with initial condition=26 for the class 2. The goal is to train two separate HMMs with the above parameters for the class 1 and the class 2 and make the prediction of the final mastery level  and compare it to the class label. 

For instance consider the following examples in Tables \ref{TB4}, \ref{TB5} and \ref{TB6}. 

$\textit{Class 1}$: Table \ref{TB4} shows an example of a game trajectory for a student who finishes four levels of the game and obtained 1.5 out of 8 in the posttest. For class 1 students, 1 in the ``State Sequence'' column indicates the lowest mastery level and 3 indicates the highest mastery level. Since there are 3 states, score in the [2,3] is mapped to label 2 and scores in the [1,2) is mapped to label 1. The following cases illustrate how final mastery level is calculated using the methods discussed in the last section.
\begin{enumerate}
	\item \textbf{Naive}: $\hat{S}= s_4 = 1$. 
    This method ignores all the past states and makes the predicted value to be the most recent state. 
    \item \textbf{Average}: $\hat{S}= \frac{1+1+1+1}{4}= 1$. This method assigns equal weight to each state and could be the best prediction method for the game since levels of the game are independent of each other and should be treated separately. 
    \item \textbf{Mode}: $\hat{S}=1$. This method predicts the final value to be the state that is repeated the most. 
\end{enumerate} 

Table \ref{TB5} shows another example of comparison between the posttest and the state trajectories for the class 1 students. The final prediction for this student is done as follows:
\begin{enumerate}
	\item \textbf{Naive}: $\hat{S}= s_{42} = 1$. 
	This method predicts the class label correctly but it ignores all the past states and does not take any past performance into the account. 
	\item \textbf{Average}: $\frac{\sum_{i=1}^{i=42}s_i}{42}=1.81 $. Since $1.81$ is less than 2 therefore the predicted label would be 1. 
	\item \textbf{Mode}: $\hat{S}=1$. The predicted label using this method is also 1, since 1 is repeated more than any other states. 
\end{enumerate} 

\begin{table}[!t]
	\caption{BIC for Class 1 in Case I}
	\vspace{-1em}
	\begin{center}
		\begin{tabular}{c|c|c|c} 
			\hline
			Row Number   & Initial Condition & BIC   & Number of States \\  
			\hline
			$1$ & 1  & 21885 & 2  \\ 
			\hline
			$2$ & 26 & 21720 & 3 \\
			\hline
			$3$ & 35 & 21636 & 3 \\
			\hline
			$4$ & 55  & 21709 & 3  \\
			\hline
			$5$ & 64 & 21779 & 4 \\  
			\hline
			$6$ & 100 & 21880 & 2 \\
			\label{TB2}	
		\end{tabular}
		
	\end{center}

\end{table}  

\begin{table}[!t]
	\caption{BIC for Class 2 in Case I}
	\vspace{-1em}
	\begin{center}
		\begin{tabular}{c|c|c|c} 
			\hline
			Row Number   & Initial Condition & BIC   & Number of States \\  
			\hline
			$1$ & 1  & 17716 & 2  \\ 
			\hline
			$2$ & 26 & 17538 & 2 \\
			\hline
			$3$ & 35 & 17541 & 4 \\
			\hline
			$4$ & 55  & 17557 & 3  \\
			\hline
			$5$ & 64 & 17631 & 4 \\  
			\hline
			$6$ & 100 & 17726 & 2 \\
			\label{TB3}	
		\end{tabular}
	\end{center}
	
\end{table}

\begin{table}[!t]
	\caption{HMM Trajectory for class 1 student in Case I}
	\vspace{-1em}
	\begin{center}
		\begin{tabular}{c|c|c} 
			\hline
			ID  & Posttest & State Sequence  \\  
			\hline
			1994 & $1.5$ & $1 1 1 1$ \\ 
			\label{TB4}	
		\end{tabular}
	\end{center}
\end{table}

\begin{table}[!t]
	\caption{HMM Trajectory for class 1 student in Case I}	
	\vspace{-1em}
	\begin{center}
		\begin{tabular}{c|c|c} 
			\hline
			ID  & Posttest & State Sequence  \\  
			\hline
			1764 & $2.5$ & $1 1 1 1 1 1 1 1 1 1 1 1 1 1 1 1 2 3 3 3 3 3$ \\ 
			~    &   ~   & $ 3 3 3 3 3 3 3 3 3 3 3 2 1 1 1 1 1 1 1 1$
			\label{TB5}
		\end{tabular}
	\end{center}
\end{table} 

$\textit{Class 2}$:
Table \ref{TB6} shows an example of a student with ID 1627 in class 2 who scored 6.17 out of 8 in the posttest and completed 49 levels of the SP game. Scores in the [1.5,2] are mapped to label 2 and scores in [1,1.5) are mapped to label 1. The final prediction for this student is as follows: 
\begin{enumerate}
	\item \textbf{Naive}: $\hat{S}= s_{49} = 1$. 
	Since $ 1<1.5$ the predicted label is 1. This is an example of forecasting error since only the last state is used for the prediction.   
	\item \textbf{Average}: $\frac{\sum_{i=1}^{i=49}s_i}{49}=1.88 $. Since $1.88$ is greater than 1.5 the predicted label would be 2. 
	\item \textbf{Mode}: $\hat{S}=2$. Since $2>1.5$ the predicted label is 2. 
\end{enumerate} 

\begin{table}[!t]
	\caption{HMM Trajectory for class 2 student in Case I}	
	\vspace{-1em}
	\begin{center}
		\begin{tabular}{c|c|c} 
			\hline
			ID    & Posttest & State Sequence  \\  
			\hline
			1564  & $6.5$ & $2 2 2 2 1 2 2 2 2 2 2 2 2 2 2 1 2 2 2 2 2 2 2 2 2 2 2 2$\\ ~       &   ~   &  $2 2 2 2 2 2 2 2 2 1 2 2 2 2 2 1 2 1 2 2 1$ 
			\label{TB6}
		\end{tabular}
	\end{center}
\end{table}

Table \ref{TB7} summarizes the accuracy for the various methods discussed in the last section by comparing the class label to the predicted value from the state trajectory. According to Table \ref{TB7} the best prediction accuracy for class 1 is for the naive method and the best prediction accuracy for class 2 is for the average and the mode methods. Among all the prediction methods described in the last section the average method is the most reliable one; this is because the naive method only accounts for the most recent mastery level and ignores all the past values. This cannot be a reliable method for the prediction using a game since different levels of the game have different game mechanics and difficulties. Therefore, every level should make a contribution to the final prediction. The average method is also more informative and provides more detail than the mode method. To better understand this consider the following cases for two different students who each finish five levels of the game:
\begin{itemize}
	\item Student A trajectory is $ 1 1 2 2 2 $. 
	\item Student B trajectory is $ 2 2 2 2 2 $
\end{itemize} 

For both students A and B the predicted label is 2 using both the mode and the average methods. However for the student A the score using the average method is 1.6 while for student B the score using the average method is 2. This shows that student B has a better performance that student A in the game. This information cannot be gained using the mode method since in both cases the state 2 is repeated the most. 

\begin{table}[!t]
	\caption{Prediction Accuracies for various Methods in Case I}
	\vspace{-1em}
	\begin{center}
		\begin{tabular}{c|c|c} 
			\hline
			Method   & Class I Accuracy & Class II Accuracy  \\  
			\hline
			$Naive$ & 97.48\% & 86.09\% \\ 
			\hline
			$Average$ & 86.55\% & 100\%  \\
			\hline
			$Mode$    & 86.55\% & 100\%  \\
			\label{TB7}	
		\end{tabular}
	\end{center}
\end{table}

$\textbf{Case II}$: Total number of moves per level is used as the observations. 

The observations are discretized to four levels according to the following rule: An expert plays all the levels of the game and the total number of moves to finish each level is recorded, then there is an $\alpha=1.5$ which can be called as a compensation factor which compensates for the game mechanics and the game difficulties. The compensation factor is multiplied by total moves per level for the expert and the observations are discretized according to this rule per level since different levels might have different difficulties. According to the Tables \ref{TB8} and \ref{TB9} the training is done by letting the number of states to be 3 for both class 1 and 2 and the initial conditions to be 35 for the class 1 and 26 for the class 2. Similar analysis to case I is done here to predict the final mastery level.

For instance consider the following examples in Tables \ref{TB10} and \ref{TB11}. 

$\textit{Class 1}$: Table \ref{TB10} shows an example of a class 1 student with ID 1768 who scored 3.88 out 8 in the posttest. The predicted mastery level using all three methods is 1. This is because 1 is repeated more than other states, the most recent state is 1 and the average value of the state trajectory is 1.85 which is less than 2. Although all three methods predict the final mastery level correctly, the average method is more informative since it provides more detail that the given student has done well on some levels since his score is close to the boundary between the class 1 and the class 2.

$\textit{Class 2}$: Table \ref{TB11} shows another example where the naive method can make an incorrect prediction. The mode or average methods predict the final mastery level to be 2 while the naive method predicts the final value to be 1 since the most recent state is 1.  
\begin{table}[!t]
	\caption{BIC for Class 1 in Case II}
	\vspace{-1em}
	\begin{center}
		\begin{tabular}{c|c|c|c} 
			\hline
			Row Number   & Initial Condition & BIC   & Number of State \\  
			\hline
			$1$ & 1  & 25895  & 2  \\ 
			\hline
			$2$ & 26 & 25850  & 7 \\
			\hline
			$3$ & 35 & 25596   & 3 \\
			\hline
			$4$ & 55  & 25613   & 3  \\
			\hline
			$5$ & 64 & 25682   & 4 \\  
			\hline
			$6$ & 100 & 25872  & 2 \\
		\label{TB8}	
		\end{tabular}
	\end{center}

\end{table}

\begin{table}[!t]
	\caption{BIC for Class 2 in case II}
	\vspace{-1em}
	\begin{center}
		\begin{tabular}{c|c|c|c} 
			\hline
			Row Number   & Initial Condition & BIC   & Number of State \\  
			\hline
			$1$ & 1  &  24462 & 2  \\ 
			\hline
			$2$ & 26 & 24206  & 3 \\
			\hline
			$3$ & 35 & 24239  & 4 \\
			\hline
			$4$ & 55  & 24231  & 3  \\
			\hline
			$5$ & 64 & 24315  & 4 \\  
			\hline
			$6$ & 100 & 24456  & 5 \\
		\label{TB9}	
		\end{tabular}
	\end{center}

\end{table} 

\begin{table}[!t]
	\caption{HMM Trajectory for class 1 student in Case II}
	\vspace{-1em}
	\begin{center}
		\begin{tabular}{c|c|c} 
			\hline
			ID   & Posttest & State Sequence  \\  
			\hline
			1768  & $3.88 $ & $2 1 1 1 1 1 1 1 1 1 1 1 1 1 1 1 2 3 3 3 3 3 3 3 3 3 3$ \\ 
			~     &  ~      & $3 3 3 3 3 3 3 3 3 2 1 1 1 1 1 1 1 1 1 1 1$\\
		\label{TB10}	
		\end{tabular}
	\end{center}
\end{table}

\begin{table}[!t]
	\caption{HMM Trajectory for class 2 student in Case II}
	\vspace{-1em}
	\begin{center}
		\begin{tabular}{c|c|c} 
			\hline
			ID    & Posttest & State Sequence  \\  
			\hline
			1573 & $7.5$ & $2 2 2 3 3 3 3 3 3 3 3 3 3 3 3 1 2 3 3 3 3 3 3 3 1 2 3 3 3 3$\\     ~      ~  &   ~    &  $3 3 1 2 3 3 3 3 1 2 2 2 3 3 1 2 3 3 3 1 $ \\
		\label{TB11}	
		\end{tabular}
	\end{center}

\end{table} 

\begin{table}[!t]
	\caption{Prediction Accuracies for Various Methods in Case II}
	\vspace{-1em}
	\begin{center}
		\begin{tabular}{c|c|c} 
			\hline
			Method   & Class I Accuracy & Class II Accuracy  \\  
			\hline
			$Naive$ &89.92\% & 92.17\% \\ 
			\hline
			$Average$ & 75.63\% & 100\%  \\
			\hline
			$Mode$    & 76.47\% & 100\%  \\
			\label{TB12}	
		\end{tabular}
	\end{center}

\end{table}

Table \ref{TB12} presents the prediction accuracies for various algorithms for the class 1 and the class 2 students. According to Table \ref{TB12} the highest prediction accuracy for class 1 is for the naive method and for class 2 is for the average and mode methods. Similar to case I, it can be argued that the average method is better than naive and is more informative than the mode method.

\begin{table}[!t]
	\caption{Statistical Results for Average Method}
	\vspace{-1em}
	\begin{center}
		\begin{tabular}{c|c|c} 
			\hline
			      & Case I & Case II  \\  
			\hline
			$Accuracy$ & 93.16\% & 87.61\% \\ 
			\hline
			$Recall$ & Class 1: 86.55\%   &   Class 1: 75.63\%  \\
					 & Class 2: 100.0\% &   Class 2: 100.0\%  \\
			\hline
			$Precision$    & Class 1: 100.0\% &   Class 1: 100.0\%  \\
						   & Class 2: 87.79\%   &    Class 2: 79.86\%  \\
		\hline
			$F_1 Score$    & Class 1: 92.79\% &   Class 1: 86.12\%  \\
						   & Class 2: 93.50\% &    Class 2: 88.80\%  \\
		 \hline
		    $AUC Score$    & 0.8644           &      0.8818   
			\label{TB13}	
		\end{tabular}
	\end{center}

\end{table}

Other metrics that are widely used in evaluating a model are recall, precision, accuracy, $F_1$ and AUC scores. Table \ref{TB13} summarizes the results for the average method for both class 1 and class 2 students under both case I and case II. High values of accuracy, recall, precision, $F_1$ and AUC scores under both case I and II suggest that the proposed method can perform strong prediction of student mastery levels. 

One important topic that needs more attention is the confounding variables. They are defined as the variables that affect both the independent and dependent variables and if not controlled properly they might change the results of experiments. For example, transfer of knowledge between the game and the posttest is a confounding variable. Transfer of knowledge is the application of the previously learned skills in a new domain. This can be the main reason why accuracy for class 1 is lower than class 2 students since class 1 students could have a harder time connecting the game concepts to the posttest. Game mechanics and the difficulty of the different levels are other confounding variables in the SP game. However there might be other confounding variables which are hard to control and might cause the change in the exam score such as students' interest, family situation, health and many more.  Since in this study a retrospective analysis of the data was conducted, it was not possible to query such factors.

\section{conclusion}
In this paper, the HMM algorithm is used to predict the students' final mastery level given their performance in various levels of the game. It was shown that despite various confounding variables affecting the students the HMM can be used as a promising solution in educational environments to model students' actions and make the prediction throughout the game. 

The results indicate that examining time series data from the game can lead to dynamic evaluation of student mastery levels throughout the time which cannot be obtained by examining only the posttest. This can be useful to make a timely intervention and provide efficient feedback. In particular, such dynamic analysis can enable students to be guided through a sequence of concepts that build on each other, thus enabling learning at their own pace. It can also be used to target human intervention for students who are struggling. 

While for this study there was no ground truth to quantify student attainment throughout the game, the strong prediction of final mastery level shows that there is considerable promise in applying the HMM to this purpose. A focus of the future research will be on how to design the interactive game experience to enable such inferences to be of high quality.

\ifCLASSOPTIONcaptionsoff
  \newpage
\fi

\bibliographystyle{IEEEtran}
\bibliography{references}






\end{document}